\documentclass[letterpaper, 10 pt, conference]{IEEEtran}
\def\IEEEtitletopspace{18pt}
\IEEEoverridecommandlockouts
\usepackage[nolist]{acronym}
\usepackage{amsmath,amssymb,amsfonts}
\usepackage{algorithmic}
\usepackage{graphicx}
\usepackage{textcomp}
\usepackage{xcolor}
\usepackage{svg}
\usepackage{adjustbox}
\usepackage{pgfplots}
\usepgfplotslibrary{statistics}
\usepackage{pgfplotstable}
\usepackage{booktabs}
\usepackage{multirow}
\usepackage{float}
\usepackage{multicol}
\usepackage{todonotes}
\usepackage{caption}
\usepackage{subcaption}
\usepackage{siunitx}
\usepackage{hyperref}
\usepackage{pifont}
\newlength{\IEEEleft}
\usepackage[absolute,overlay]{textpos}
\usepackage[capitalise]{cleveref}
\def\BibTeX{{\rm B\kern-.05em{\sc i\kern-.025em b}\kern-.08em
    T\kern-.1667em\lower.7ex\hbox{E}\kern-.125emX}}

\usepackage{hyperref}
\hypersetup{
    colorlinks=true,
    linkcolor=black,
    citecolor=black,
    filecolor=black,
    urlcolor=black,
}

\usepackage[style=ieee,
            doi=false,
            url=false,
            mincitenames=1,
            maxcitenames=1,
            minbibnames=6,
            maxbibnames=6,
            backend=biber]{biblatex}  
\AtBeginBibliography{\footnotesize}

\begin{acronym}
	\acro{GNSS/INS}{global navigation satellite system/inertial navigation system}
\end{acronym}

\begin{document}

\title{
    Impact of Localization Errors on Label Quality for Online HD Map Construction
}

\author{
    Alexander Blumberg$^{1*}$,
    Jonas Merkert$^{1 *}$,
    Richard Fehler$^{2}$,
    Fabian Immel$^{2}$,\\
    Frank Bieder$^{2}$,   
    Jan-Hendrik Pauls$^{1}$ 
    and 
    Christoph Stiller$^{1}$
    \thanks{$^{*}$These authors contributed equally to this work.}
    \thanks{
        $^{1}$Institute of Measurement and Control Systems, Karlsruhe Institute of Technology (KIT),
        Karlsruhe, Germany
        {\tt\footnotesize \{alexander.blumberg, jonas.merkert, jan-hendrik.pauls, stiller\}@kit.edu}
    }%
    \thanks{
        $^{2}$FZI Research Center for Information Technology, Karlsruhe, Germany
        {\tt\footnotesize \{fehler, immel, bieder\}@fzi.de}
    }%
}

\maketitle

\begin{textblock*}{\textwidth}(\IEEEleft, 265mm) 
\footnotesize
© 2025 IEEE. Personal use of this material is permitted. Permission from IEEE must be obtained for all other uses, in any current or future media, including reprinting/republishing this material for advertising or promotional purposes, creating new collective works, for resale or redistribution to servers or lists, or reuse of any copyrighted component of this work in other works.
\end{textblock*}

\begin{abstract}

High-definition (HD) maps are crucial for autonomous vehicles, but their creation and maintenance is very costly.
This motivates the idea of online HD map construction.
To provide a continuous large-scale stream of training data, existing HD maps can be used as labels for onboard sensor data from consumer vehicle fleets.
However, compared to current, well curated HD map perception datasets, this fleet data suffers from localization errors, resulting in distorted map labels. 

We introduce three kinds of localization errors, Ramp, Gaussian, and Perlin noise, to examine their influence on generated map labels.
We train a variant of MapTRv2, a state-of-the-art online HD map construction model, on the Argoverse~2 dataset with various levels of localization errors and assess the degradation of model performance.
Since localization errors affect distant labels more severely, but are also less significant to driving performance, we introduce a distance-based map construction metric.
Our experiments reveal that localization noise affects the model performance significantly.
We demonstrate that errors in heading angle exert a more substantial influence than position errors, as angle errors result in a greater distortion of labels as distance to the vehicle increases. Furthermore, we can demonstrate that the model benefits from non-distorted ground truth (GT) data and that the performance decreases more than linearly with the increase in noisy data.
Our study additionally provides a qualitative evaluation of the extent to which localization errors influence the construction of HD maps.

\end{abstract}

\begin{figure}[t]
    \centering
    \adjustbox{trim=1.6cm 0cm 1.6cm 0cm, width=\columnwidth, clip}{
        \includesvg{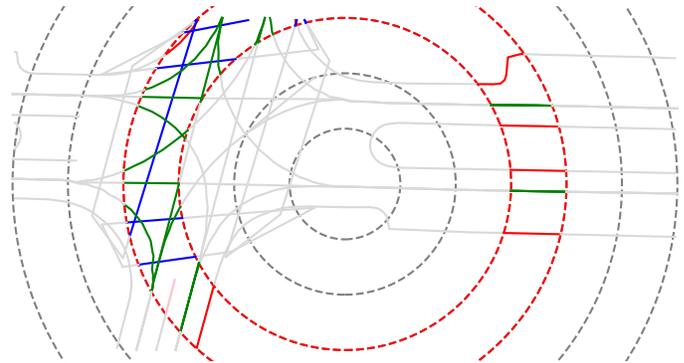}
    }
    \caption{Ring evaluation input (colored) over the entire scene (grey). Displayed are the map elements road boundaries (red), centerlines (green), pedestrian crossings (blue), and dividers (pink). Rings plotted as dashed with the currently observed ring (red) and other rings (grey).}
    \label{fig:ring_evaluation}
\end{figure}
\begin{figure*}[t]
    \centering
    \adjustbox{trim=0cm 3.2cm 0cm 0cm, width=\linewidth, clip}{
        \includesvg{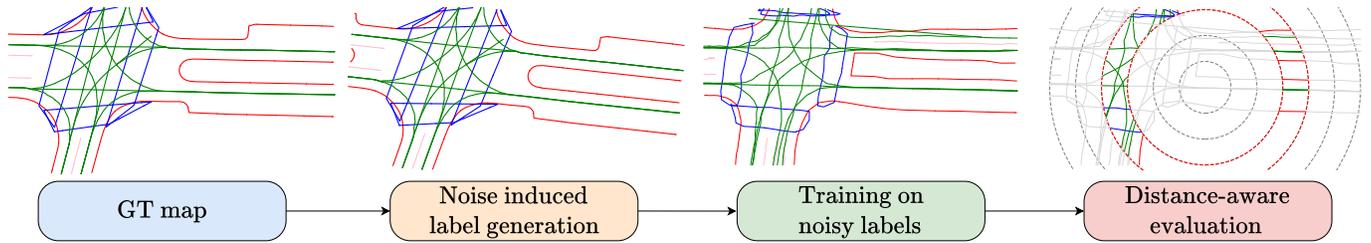}
    }
    \caption{Flowchart with an overview from map to evaluation. The noise application is described in \cref{noise_types}, details about the ground truth (GT) and training can be found in \cref{sec:setup} and we introduce our distance-aware evaluation metric in \cref{evaluation_metric}.}
    \label{fig:flowchart_overview}
\end{figure*}

\section{Introduction}

High-definition (HD) maps are a fundamental component of today's autonomous vehicle architectures. In contrast to standard-definition (SD) navigation maps, HD maps contain more detailed, more precise and additional semantic information about the topology of the road network.
However, HD map construction is done mainly with mobile mapping platforms or survey vehicles equipped with highly precise sensors and reference \ac{GNSS/INS} units. For that reason, HD map generation and maintenance is costly and time-consuming.
A novel idea is aiming to utilize machine learning methods to perceive and construct HD maps online from onboard sensor data~\cite{liao2023MapTR, liao2024maptrv2, liu2023vectormapnet, chen_maptracker_2025, wang2024StreamQueryDenoising}.

Existing HD map construction datasets like Argoverse~2~\cite{Argoverse2} or nuScenes~\cite{caesar_nuscenes} utilize sensor data with high precision but are limited in size and diversity.
While it is unclear how to scale the spatial extent of HD map coverage, existing vehicle fleets can provide a basically unlimited amount of sensor data.
This data with high diversity in sensor setups and environment conditions, can be used as training data for online HD map construction when recorded in places with HD map coverage. 
However, their ego localization is less reliable, and the resulting vehicle poses are less accurate.
This, in turn, leads to distorted HD map ground truth.
For this reason, it is important to know, how localization errors affect ground truth quality, how this impairs online HD map construction performance, and hence, which level of precision the ego localization has to reach for HD map construction models to predict sufficiently precise maps.

The contributions of this work are:
\begin{itemize}
    \item We are the first to model diverse localization noise based on real-world localization uncertainty patterns~(\cref{noise_types}).
    For each of the three distortion types, we apply increasing levels of noise and apply it to a map perception dataset. 
    \item We train a state-of-the-art online HD map construction model using the various distortions.
    This allows us to compare the degradation of model performance.
    \item Since angular errors affect distant map elements more severely, we introduce a novel distance-aware map construction metric that evaluates label distortions and the model performance within different perception ranges~(\cref{evaluation_metric}).
    \item To examine the cause and effects, we compare original ground truth, distorted labels, and model predictions using the new distance-aware metric.
\end{itemize}

\section{Related Work}
\subsection{Online HD Map Construction}
Today's state-of-the-art HD Map Construction models utilize sensor data of the surrounding scene to predict vectorized map elements in their sparse polyline representation. Therefore, features are extracted from camera or LiDAR data and transformed into a bird's-eye-view feature space. Subsequently, a Detection Transformer (DETR)~\cite{carion_2020_detr} architecture predicts the vectorized map elements.

VectorMapNet~\cite{liu2023vectormapnet} leverages a coarse-to-fine approach to predict the map elements in an autoregressive manner. In contrast, MapTR~\cite{liao2023MapTR} and MapTRv2~\cite{liao2024maptrv2} address the speed limitations of the autoregressive approach by introducing a new query design in the transformer decoder and treat the task as a point set prediction problem with a fixed number of points. 
Recent approaches have demonstrated significant improvements by taking advantage of previous predictions in a temporal or spatial manner~\cite{chen_maptracker_2025, wang2024StreamQueryDenoising, Yuan_2024_WACV}. Not only the inclusion of the temporal aspect is the subject of current research, but also the expansion of the sensor data to include additional prior information from existing SD navigation maps or older HD maps from previous drives~\cite{luo2024smerf, zeng2024drivingpriormapsunified, immel2024m3trgeneralisthdmap}.

\subsection{Effect of noisy labels and sensor deviations}

This work examines the impact of noisy labels during the training process on the outcome of online map construction. In recent years, the evaluation of label errors has garnered increased interest in the research community, and novel methods for classification and object detection tasks have been developed to handle noisy labels effectively~\cite{northcutt2021labelerrors, Li_2020_CVPR, Chen_2023_ICCV}. Adhikari et al.~\cite{adhikari_effect_2021} examine the effect of missing labels in object detection and therefore remove a percentage of bounding box labels during training. Besides missing labels, Freire et al.~\cite{freire_beyond_2024} and Zhu et al.~\cite{zhu2024robusttinyobjectdetection} also explore the impact of incorrect or imprecise labels, such as class shifts or bounding box distortions.

In addition to investigating the issue of noisy labels in object detection, research in online HD map construction also examines the impact of erroneous or poor-quality data. MultiCorrupt~\cite{Beemelmanns_iv_2024} and MapBench~\cite{hao2024mapbench} evaluate the effect of weather conditions, sensor failures, or motion blur. Conversely, MapTR~\cite{liao2023MapTR, liao2024maptrv2} explores the impact of camera deviations on map construction, such as a shift of the camera position or calibration errors. The aforementioned studies investigate the effect of noise on the input and validation data. In contrast, our research focuses on the impact of noisy map labels during training, arising from localization and heading errors, on the performance of the model. The related work section listed above shows that there are no comparable studies to date.

\begin{figure*}[h]
    \centering
    \adjustbox{trim=5cm 3cm 5cm 0.75cm, width=\linewidth, clip}{
        \includesvg{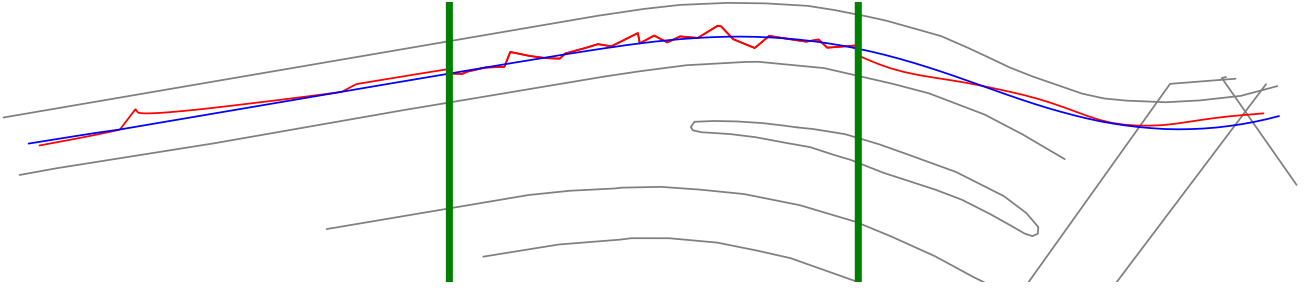}
    }
    \caption{Qualitative example with ground truth (blue) and noisy localization (red) with Ramp noise (left), Gaussian noise (middle) and Perlin noise (right).}
    \label{fig:noised_trajectories}
\end{figure*}


\section{Methodology}
\subsection{Noise types}
\label{sec:noise_types}
We include three different noise cases Ramp, Gaussian and Perlin noise, for which we each calculate an altered position $x_{\eta}(t)$, $y_{\eta}(t)$ and orientation $\theta_{\eta}(t)$ based on the original pose $x_{gt}(t)$, $y_{gt}(t)$ and $\theta_{gt}(t)$.
\label{noise_types}
\subsubsection{Ramp noise}
A vehicle's localization module may fuse input from various sources such as GNSS, IMU, magnetometer, wheel odometry, visual localization, etc. Based on a vehicle dynamic model such, as the single-track model~\cite{Riekert1940ZurFD} and a filter such as an Extended Kalman filter this module will return a combined localization output. This allows for limiting the downsides of certain localization means such as higher noise levels for GNSS localization of drifting properties for IMU and wheel odometry-based approaches.

Effects such as blockage of the GNSS signal, whether it is partly through buildings or foliage or totally through tunnels or bridges, as well as GNSS multipath effects~\cite{Kos2010EffectsOM} may however limit the availability or at least the precision and confidence of a GNSS signal. If no other absolute means of localization is available at this point, the calculated vehicle pose will drift away from its actual position. In case a precise GNSS lock with high confidence returns, the calculated vehicle pose will jump back to its actual position in a short time.

We model this behavior with a Ramp function $R(t)_n$ that generates a uniformly distributed interval between 4s and 10s and interpolates in this interval between no offset and a uniformly distributed translation offset $T_n$ between 0 and $\epsilon_L$ as well as a uniformly distributed angular offset $\theta_n$ between $-\epsilon_R$ and $\epsilon_R$ around the yaw axis. The direction $\alpha_n$ of the translation offset is uniformly distributed between 0 and 360\textdegree. The Ramp function is called again as soon as the previous interval is exceeded:
\begin{gather}
    \begin{pmatrix}x_{\eta}(t) \\ y_{\eta} (t)\\ \theta_{\eta}(t)\end{pmatrix}=\begin{pmatrix}x_{gt}(t) \\ y_{gt}(t) \\ \theta_{gt}(t)\end{pmatrix}+\begin{pmatrix}cos(\alpha_n)T_{n} \\ sin(\alpha_n)T_{n} \\ \theta_{n}\end{pmatrix}R_n(t)\\
    R_n(t)=\frac{t-t_n}{t_{n+1}-t_n}; t_{n+1}> t\geq t_n \\
    t_{n+1}\sim \mathcal{U}(t_n+4s,t_n+10s);\alpha_n\sim \mathcal{U}(0,360^\circ)\\ 
    T_{n}\sim \mathcal{U}(0,\epsilon_L); \theta_{n}\sim \mathcal{U}(-\epsilon_R,\epsilon_R)\text{.}
\end{gather}
\cref{fig:noised_trajectories} shows an example of the noisy localization on the left.

\begin{table*}[t]
\centering
\caption{Comparison between prediction results for the noise types Ramp, Gaussian, and Perlin evaluated using Average Precision (AP). As noise parameters, we defined the maximum translation error~$\epsilon_L$, maximum angular error~$\epsilon_R$, for the Gaussian noise case standard deviation in translation~$\sigma_L$ and angular~$\sigma_R$ direction, noise ratio~$NR$ as the percentage of altered ground truth, and heading correction~$HC$ as correction to the now altered driving direction.}
\begin{tabular*}{\textwidth}{@{\extracolsep{\fill}}lrrrrrlllllll@{}}
\toprule
\multirow{2}{*}{\textbf{Experiment}} & \multicolumn{6}{c}{\textbf{Noise parameters}} & \multirow{2}{*}{$\mathbf{AP_{dsh}}$} & \multirow{2}{*}{$\mathbf{AP_{sol}}$} & \multirow{2}{*}{$\mathbf{AP_{bou}}$} & \multirow{2}{*}{$\mathbf{AP_{cen}}$} & \multirow{2}{*}{$\mathbf{AP_{ped}}$} & \multirow{2}{*}{\textbf{mAP}} \\ 
\cmidrule{2-7}
& $\epsilon_L$ & $\sigma_L$ & $\epsilon_R$ & $\sigma_R$ & $NR$ & $HC$ & & & & & & \\ 
\midrule
Baseline $B$ & - & - & - & - & - & - & 36.7 & 54.4 & 49.6 & 48.3 & 40.0 & 45.8 \\
Ramp $R_1$ & 2 m & - & 1° & - & 50 \% & \ding{51} & 34.8 & 52.1 & 46.0 & 46.1 & 36.5 & 43.1 \\
Ramp $R_2$ & 2 m & - & 0° & - & 100 \% & \ding{55} & 32.0 & 49.0 & 46.5 & 44.3 & 33.6 & 41.1 \\

Ramp $R_3$ & 2 m & - & 0° & - & 100 \% & \ding{51} & 30.6 & 46.4 & 43.7 & 42.7 & 33.1 & 39.3 \\
Ramp $R_4$ & 2 m & - & 1° & - & 100 \% & \ding{51} & 29.8 & 45.6 & 44 & 43.1 & 31.8 & 38.9 \\
Ramp $R_5$ & 1 m & - & 0.5° & - & 100 \% & \ding{51} & 33.9 & 50.7 & 46.0 & 46.6 & 37.5 & 43.0 \\
Gaussian $G_1$ & 2 m & 0.5 m & 0° & 0° & 100 \% & \ding{55} & 33.4 & 52.4 & 46.5 & 40.4 & 37.6 & 42.1 \\
Gaussian $G_2$ & 2 m & 0.5 m & 0.3° & 1° & 100 \% & \ding{55} & 33.9 & 53.3 & 47.4 & 44.7 & 37.2 & 43.3 \\
Gaussian $G_3$ & 2 m & 0.5 m & 3° & 10° & 100 \% & \ding{55} & 25.2 & 36.4 & 41.8 & 31.2 & 35.4 & 34.0 \\
Perlin $P_1$ & 0 m & - & 0.5° & - & 100 \% & \ding{55} & 38.3 & 55.4 & 47.7 & 48.1 & 39.2 & 45.7 \\
Perlin $P_2$ & 1 m & - & 0° & - & 100 \% & \ding{51} & 29.4 & 42.8 & 42.0 & 39.8 & 34.9 & 37.8 \\
Perlin $P_3$ & 2 m & - & 0° & - & 100 \% & \ding{51} & 17.2 & 29.5 & 31.9 & 25.6 & 30.8 & 27.0 \\
Perlin $P_4$ & 2 m & - & 0.5° & - & 100 \% & \ding{51} & 17.6 & 31.5 & 32.5 & 27.2 & 30.5 & 27.9 \\
\bottomrule
\end{tabular*}
\label{table_map}
\end{table*}

\begin{figure*}[t]
    \centering
    \begin{minipage}[b]{0.9\linewidth}
        \begin{minipage}[b]{.66\linewidth}
        \centering
        \includegraphics[width=\linewidth]{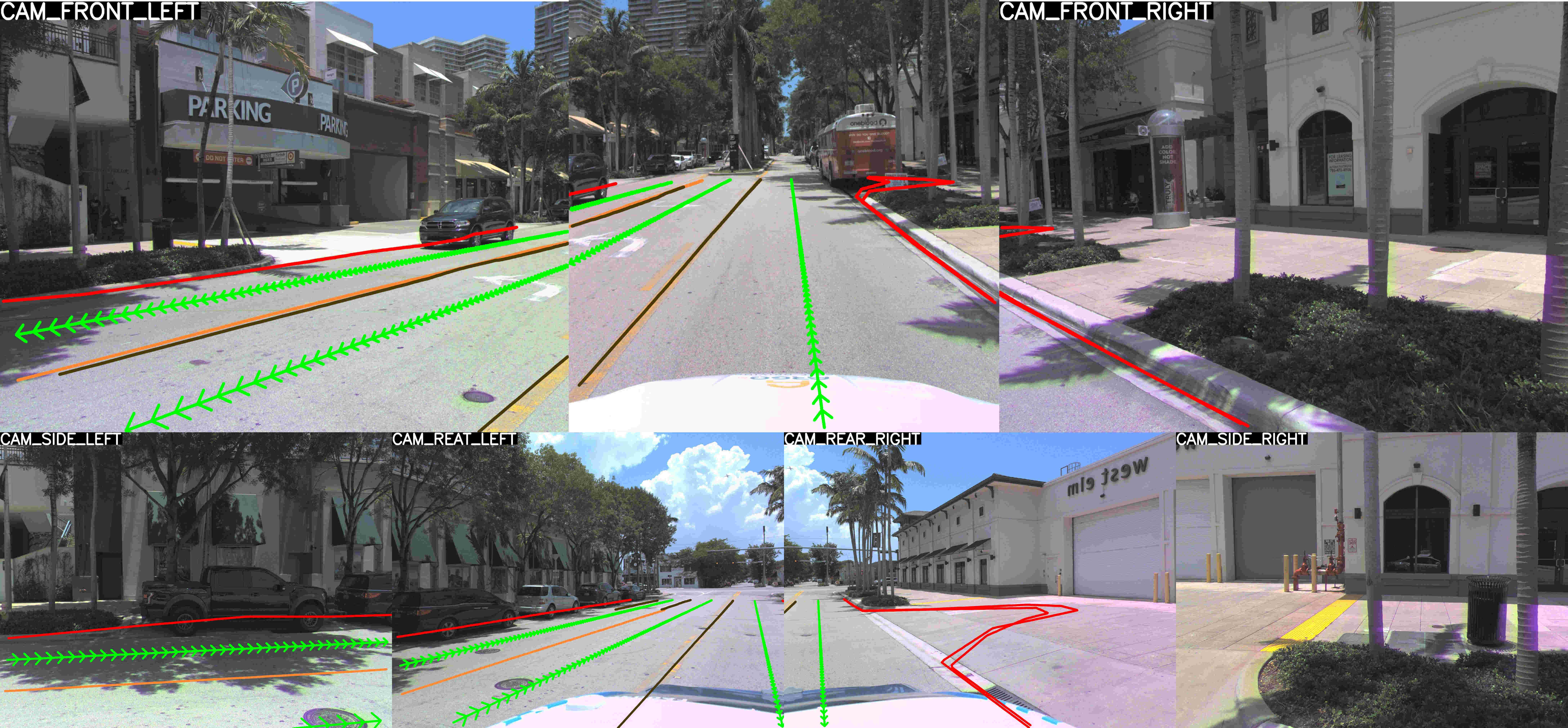}
        \label{fig:image1}
        \end{minipage}
        \hfill
        \begin{minipage}[b]{.33\linewidth}
            \centering
            \includegraphics[width=\linewidth]{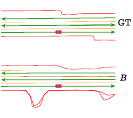}
            \label{fig:image1}
        \end{minipage}
        \\
        \begin{minipage}[b]{\linewidth}
            \centering
            \includegraphics[width=\linewidth]{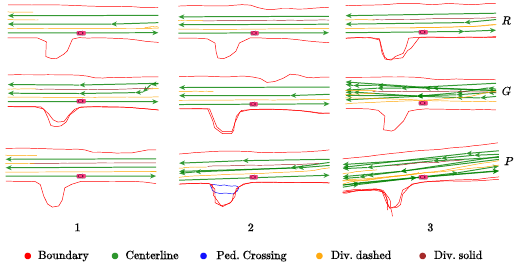}
            \label{fig:image3}
        \end{minipage}
    \end{minipage}
    \caption{Qualitative prediction results for the Baseline~$B$ (trained without noisy GT labels), Ramp~$R_1-R_3$, Gaussian~$G_1-G_3$, and Perlin~$P_1-P_3$ for the camera input (top left, Baseline results are projected into the image) and the ground truth map~$GT$.}
    \label{fig:prediction_example}
\end{figure*}
\begin{figure*}[t]
    \centering
    \adjustbox{trim=0cm 0.2cm 0cm 0.3cm, width=\linewidth, clip}{
        \includesvg{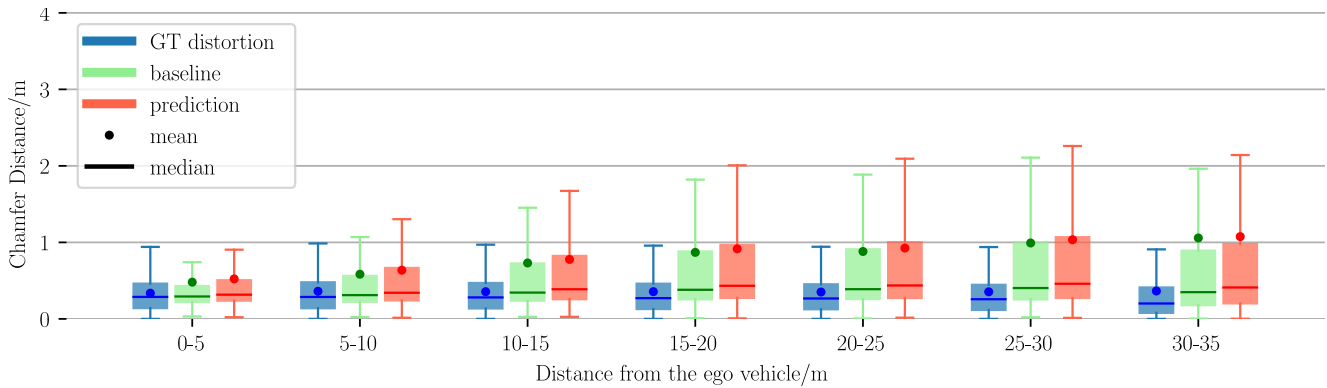}
    }
    \caption{Gaussian $G_1$ (GT distortion and prediction) and Baseline $B_1$}
    \label{fig:g1}
\end{figure*}
\begin{figure*}[h]
    \centering
    \adjustbox{trim=0cm 0.2cm 0cm 0.3cm, width=\linewidth, clip}{
        \includesvg{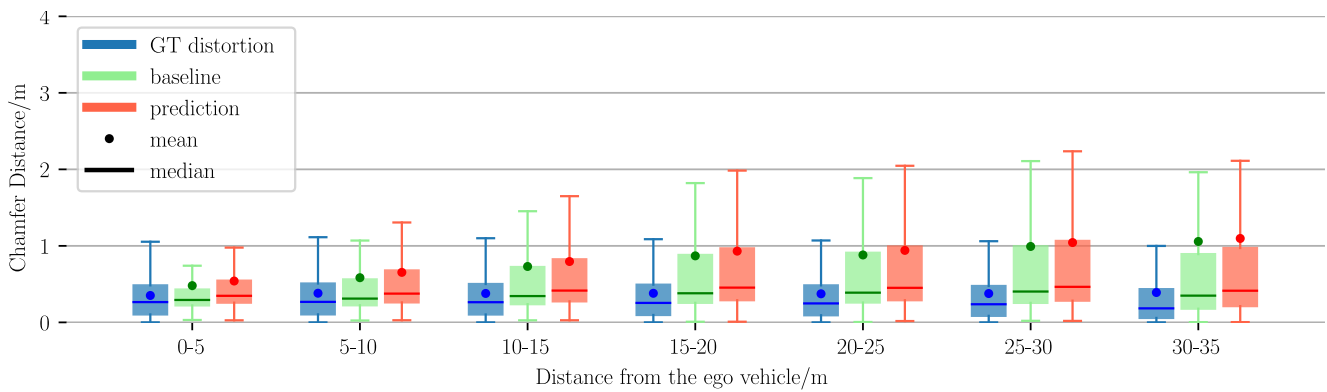}
    }
    \caption{Ramp $R_2$ (GT distortion and prediction) and Baseline $B_1$}
    \label{fig:r2}
\end{figure*}

\subsubsection{Gaussian noise}
One of the simplest ways of obtaining a localization is using the output from an absolute localization source such as GNSS without filtering. This would provide us with a somewhat noisy output around the vehicle's actual position.

We model this by applying a truncated Gaussian distribution $\mathcal{N}_{[\ ]}$ with a mean of 0 as an offset for both the translation $T_t$ and the angular case $\theta_t$. The direction of the translation offset is uniformly distributed between 0 and 180\textdegree:
\begin{gather}
    \begin{pmatrix}x_{\eta}(t) \\ y_{\eta} (t)\\ \theta_{\eta}(t)\end{pmatrix}=\begin{pmatrix}x_{gt}(t) \\ y_{gt}(t) \\ \theta_{gt}(t)\end{pmatrix}+\begin{pmatrix}cos(\alpha_n)T_{t} \\ sin(\alpha_n)T_{t} \\ \theta_{t}\end{pmatrix}\\
    T_t \sim \mathcal{N}_{[\ ]}(\mu=0, \sigma_{L}^2, -\epsilon_L, \epsilon_L)\\
    \theta_t \sim \mathcal{N}_{[\ ]}(\mu=0, \sigma_{R}^2, -\epsilon_R, \epsilon_R)\\
    \alpha_t\sim \mathcal{U}(\SI{0}{\degree},\SI{180}{\degree})\text{.}
\end{gather}

\cref{fig:noised_trajectories} shows an example of the noisy localization in the middle.
\subsubsection{Perlin noise}

In addition to blocked or raw GNSS signals, we can also obtain a stable but poor GNSS signal with some noise in it. This results in a smooth, wavelike path, as opposed to a GNSS drift or zigzagging GPS trajectory. This phenomenon can, for instance, be observed in the context of badly adjusted tracking in case an improperly adjusted Kalman filter is employed.

To model this behavior, we apply Perlin noise~\cite{perlin_noise_2002} to the trajectory points and the vehicle heading. Therefore, we model the noise for longitude and lateral coordinate and the vehicle heading each with a random generator to generate different Perlin noise functions.
To control the structure of the noise we use two parameters. The first one is the $octave$ parameter, which specifies the number of rectangles in each [0,1] range. This parameter is employed to define the frequency of the noise and, consequently, its degree of fineness. Given that the sampling of the points, and thereby the smoothness of the applied Perlin noise, is influenced by the vehicle's speed and GNSS sampling, we introduce the step length parameter, denoted by $\gamma$. By multiplying the $x$- and $y$-coordinates of the trajectory with the step length parameter~$\gamma$ we can vary the smoothness of the noise:

\begin{equation}
    \begin{pmatrix}x_{\eta}(t) \\ y_{\eta} (t)\\ \theta_{\eta}(t)\end{pmatrix}=\begin{pmatrix}x_{gt}(t) \\ y_{gt}(t) \\ \theta_{gt}(t)\end{pmatrix}+\begin{pmatrix}P_x(x_t \cdot \gamma, y_t \cdot \gamma)  \\ P_y(x_t \cdot \gamma, y_t \cdot \gamma) \\ P_\theta(x_t \cdot \gamma, y_t \cdot \gamma)\end{pmatrix}\text{.}
\end{equation}

In our experiments, we set $\gamma$ to 1000 and $octave$ to 10 to obtain smooth wavering trajectories. The result of a trajectory with Perlin noise is shown on the right side in \cref{fig:noised_trajectories}.

For the parameter selection of all noise types, we plotted noisy trajectories with different parameter sets on aerial images and selected those parameter sets most fitting to our experience with localization error patterns.

\subsubsection{Heading correction}
\label{heading_correction}
A resulting heading direction that is majorly different from the current direction of travel would contradict common vehicle dynamic models such as the linear one-track model. We therefore include the option to correct the vehicle heading accordingly:
\begin{equation}
    \theta_{\text{noise}} = \theta_{\text{gt}}+\angle \Delta\mathbf{p}_{\text{gt}}\Delta \mathbf{p}_{\text{noise}}
\end{equation}

Since the Gaussian noise case is aimed to explore the usage of raw sensor data for pose acquiration in contrast to a more sophisticated model-based filtering approach, the effects of the heading correction are only investigated for Ramp and Perlin noise cases.

\subsubsection{Partly altered training data}
In light of the aforementioned factors, localization noise may only be present in a limited section of the data at hand. It is therefore of interest to investigate the influence of only partially altered training data on the model's performance. This is modeled by applying the previously introduced noise types to only a percentage $\mathbf{NR}$ of the ground truth while leaving the remaining unaltered.
\subsection{Distance-aware map construction metric}
\label{evaluation_metric}
MapTR~\cite{liao2023MapTR, liao2024maptrv2}, VectorMapNet \cite{liu2023vectormapnet} and M3TR \cite{immel2024m3trgeneralisthdmap} use the average precision (AP) as an evaluation metric and a perception range of [-15m,15m] for the X-axis and [-30m,30m] for the Y-axis. For the calculation of the AP, the chamfer distance with the thresholds $T = \{0.5m, 1.0m, 1.5m\}$ is used for matching between prediction and ground truth. Then, the final AP is calculated as the average across all thresholds. A problem of this evaluation metric is that spatial relations between the vehicle and the predicted object are ignored. From the perspective of a vehicle, it is more challenging to predict the precise location of objects at a considerable distance than those in the immediate vicinity. Furthermore, in the autonomous driving task, those objects that are closer to the vehicle have a higher priority to be predicted correctly and precisely, because objects that are farther away can be predicted more precisely in future time steps when they get closer to the vehicle.

To address this gap of missing spatial relations between vehicle and prediction, we introduce a new distance-aware evaluation metric for HD map construction. Analog to~\cite{liao2024maptrv2, liao2023MapTR, liu2023vectormapnet, chen_maptracker_2025} the chamfer distance is used to assign the prediction to the ground truth elements.

After that, annuli~$\mathcal{D}$, from now on referred to as rings,  with equal width are calculated centered around the actual position of the vehicle. We process each ring individually. For each ring, we iterate over all assigned prediction and ground truth element pairs. At this point, we upsample these elements linearly with a maximum point-to-point distance of 1m. Now we differentiate:
\begin{itemize}
    \item Element pairs, for which at least one element does not include points with overlap with the ring's area are discarded at this point.
    \item For element pairs that only consist of points overlapping with the ring's area the chamfer distance between the element pair is calculated and added to the results list.
    \item For element pairs that consist of both points overlapping with the ring and points outside the ring, polylines are interpolated linearly at ring borders to include a point right at the ring borders. We now discard all points that do not overlap with the ring area and separate polylines into multiple polylines at ring borders.
    
    Since each element (ground truth or prediction) is now represented by multiple sub-elements, we need to calculate an assignment between ground truth and prediction sub-elements within this subset. We again use chamfer distance for this assignment. At this point, we set an upper bound with the radius of the ring's outer circle for sub-elements to be assigned. We add the value of the chamfer distance between the assigned sub-element pairs to the results list. 
\end{itemize}
\cref{fig:ring_evaluation} shows accepted (sub-) elements in color with rejected polylines in gray for an example ring and ground truth sample. The results list entries are best displayed in a statistical representation such as a box plot.

The formula for the distance-aware chamfer distance $\mathrm{CD}_\circledcirc$ for a GT element~$\mathcal{S}_1$, the assigned prediction~$\mathcal{S}_2$, and the distance ring~$\mathcal{D}$ is written in \cref{eq:directed_chamfer} and \cref{eq:chamfer}.

\begin{equation}
\overline{\mathrm{CD}}_{\circledcirc}\left(\mathcal{S}_1, \mathcal{S}_2, \mathcal{D}\right)=\frac{1}{\mathcal{S}_1} \sum_{p \in \mathcal{S}_1 \cap D} \min _{q \in \mathcal{S}_2 \cap D}\|p-q\|_2
\label{eq:directed_chamfer}
\end{equation}

\begin{equation}
\mathrm{CD}_\circledcirc\left(\mathcal{S}_1, \mathcal{S}_2, \mathcal{D}\right)=\frac{\overline{\mathrm{CD}}_{\circledcirc}\left(\mathcal{S}_1, \mathcal{S}_2, \mathcal{D}\right)+\overline{\mathrm{CD}}_{\circledcirc}\left(\mathcal{S}_2, \mathcal{S}_1, \mathcal{D}\right)}{2}
\label{eq:chamfer}
\end{equation}

This metric definition allows a real comparison of the precision of the predicted elements in terms of a geometric distance, ignoring multiple or missing predictions, which are covered in the AP metric.
\section{Evaluation}
\subsection{Experimental setup}
\label{sec:setup}

We conduct our experiments on the Argoverse 2 dataset~\cite{Argoverse2}. To avoid geographical overlap between train, validation, and test sets, we implement the geographical split proposed by Lilia et al.~\cite{Lilja_2024_CVPR}. As the image backbone, we use ResNet50~\cite{resnet} and the hyperparameters and experimental setup mostly from Vanilla MapTRv2~\cite{liao2024maptrv2} for all models, whereby we follow \cite{immel2024m3trgeneralisthdmap} for the geographical split and prediction of different divider types.

We apply the noise types introduced in \cref{noise_types} only to the training split at label generation time. The labels generated therefore contain a shifted and/or turned local "cutout" of the global map. At training time, the current model only "sees" the same set of altered labels. The test and validation split remain unaltered.

Our focus lies on exploring a broader set of parameters instead of reaching full convergence. Therefore, we find a training over a total of 12 epochs sufficient, also limiting the overall computational requirements.  We evaluate using the AP in \cref{table_map} as well as our newly defined distance-aware evaluation metric.

We present the effect of noise applied to the train split and evaluate the respective model's performance against the validation ground truth in \cref{fig:g1}, \cref{fig:r2}, \cref{fig:p2} and \cref{fig:p3} as a selection of our performed experiments.

Experiments and their relevant noise parameters can be referenced to each other through their annotation, as defined in the first column of \cref{table_map}. We also examined the Precision during the assignment for all experiments, but there were no substantial changes regarding the Baseline and it remains stable at 0.3.

\subsection{Qualitative analysis}
To present prediction artifacts typical to specific noise scenarios we include \cref{fig:prediction_example} as a comparison between the Ground Truth~$GT$, Baseline~$B$, Ramp~$R_{1-3}$, Gaussian~$G_{1-3}$ and Perlin~$P_{1-3}$.

Slight label errors as shown in Ramp~$R_1$, Gaussian~$G_1$, and Perlin~$P_1$ produce comparable predictions to the Baseline~$B$ model. Medium levels of label errors like in Ramp~$R_2$, Gaussian~$G_2$, and Perlin~$P_2$ will still manage to maintain the overall structure and therefore the logic behind the scene albeit with increased levels of warping. Additionally, the Perlin~$P_2$ model starts to get confused by the noise and predicts wrong map elements, such as the pedestrian crossing or double centerline detections at the edge of the perception range.

With increased label error, models tend to output a higher amount of multiple predictions for the same map element, beginning with centerline predictions, see Gaussian~$G_3$. At even higher label errors, multiple predictions for other object classes occur more often and the entire scene loses its overall structure and orientation, see Perlin~$P_3$.

\subsection{Temporal aspect}
While all of our noise cases are motivated through scenarios that can be found in real-world applications, they can in general be divided into two categories. While the Perlin and Ramp case (within the Ramp interval) are temporally consistent, the Gaussian case is not. 

When comparing $R_2$ and $G_1$, which share comparable levels of label distortion according to \cref{fig:g1} and \cref{fig:r2}, the resulting prediction evaluation turns out not majorly different according to both the mAP metric with 41.1 and 42.1 as well as our distance aware metric.

Given that the architecture of MapTRv2 \cite{liao2024maptrv2} does not incorporate a temporal aspect, this behavior matches the expectations. We would expect a model considering the temporal aspect such as \cite{chen_maptracker_2025} or \cite{wang2024StreamQueryDenoising} to provide better performance for Ramp or Perlin noise for the same level of distortion.

\begin{figure*}[t]
    \centering
    \adjustbox{trim=0cm 0.2cm 0cm 0.3cm, width=\linewidth, clip}{
        \includesvg{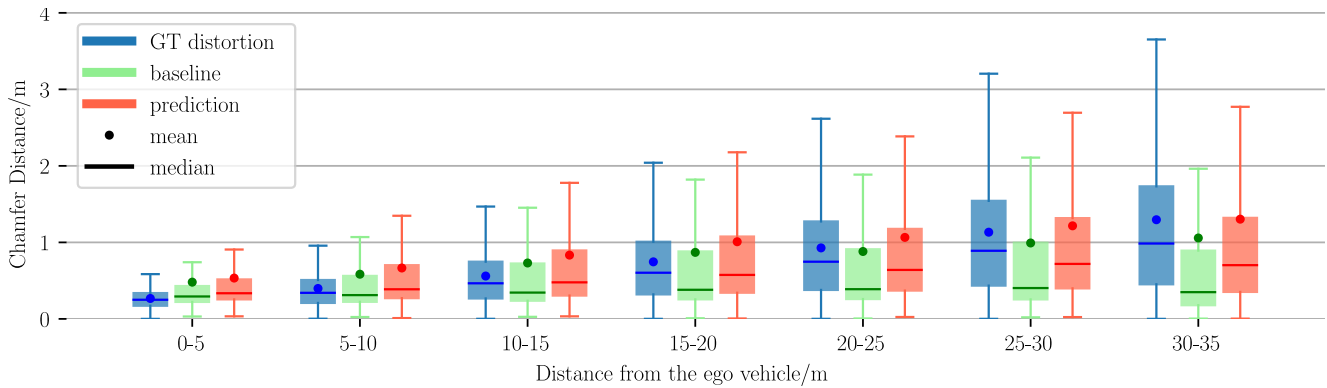}
    }
    \caption{Perlin $P_2$ (GT distortion and prediction) and Baseline $B_1$}
    \label{fig:p2}
\end{figure*}
\begin{figure*}[h]
    \centering
    \adjustbox{trim=0cm 0.2cm 0cm 0.3cm, width=\linewidth, clip}{
        \includesvg{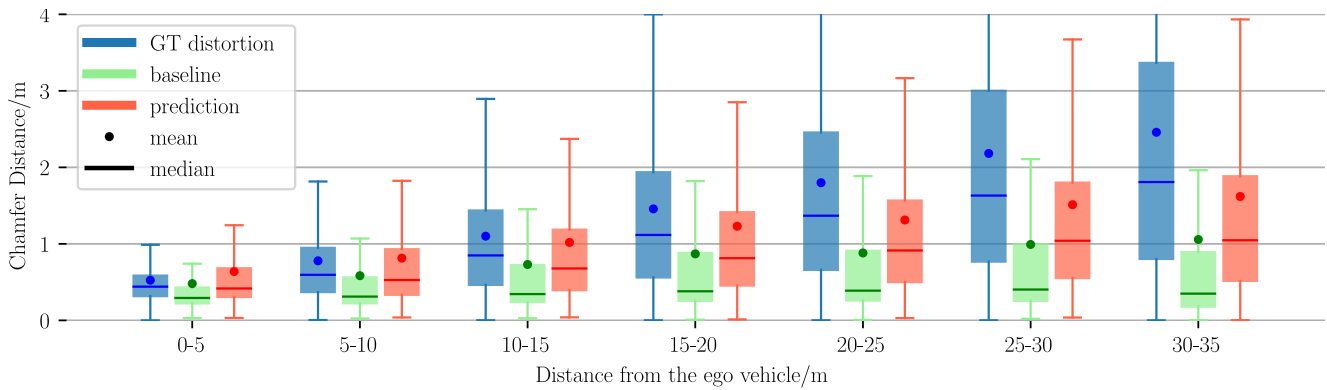}
    }
    \caption{Perlin $P_3$ (GT distortion and prediction) and Baseline $B_1$}
    \label{fig:p3}
\end{figure*}

\subsection{Partly altered training data}
In terms of only partially altered training data, the experiment Ramp~$R_1$ is conducted as illustrated in \cref{table_map}. In comparison to Ramp~$R_4$ and Baseline~$B$, concerning the percentage of data altered, Ramp~$R_1$ demonstrates a disproportionately superior performance in terms of mAP, exhibiting comparable performance to the Baseline $B$. In contrast, Ramp~$R_4$ demonstrates a greater performance degradation, exceeding twofold, when the noise ratio is doubled from 50\% to 100\%. This finding indicates that the model's performance is enhanced by the presence of correct labels and its capacity to adapt to noisy labels is augmented when correct labels are available in the dataset.

\subsection{Influence of heading correction}
A special focus of this work lies in the difference in applying a purely translational offset in comparison to an angular error.

Throughout this work, we recognized that including a heading correction~$HC$ using a pragmatic noise parameter set as shown in \cref{table_map} has a greater effect on both the imposed error through the altered training label at an increased distance from the vehicle and the resulting model's performance in terms of mAP. 

When analyzing the induced error for the training of Perlin~$P_2$ and $P_3$, as shown in \cref{fig:p2} and \cref{fig:p3}, both result in an almost linear relationship between mean/median error in chamfer distance and distance from ego pose, a relationship expected from a predominantly angular error. Additionally, the error is effectively twofold with a doubling in $\epsilon_L$. Relative to this, the error induced in Ramp~$R_2$ and Gaussian~$G_1$ as shown in \cref{fig:g1} and \cref{fig:r2} respectively as a purely translational error is considerably less severe. 

Focusing on the resulting model's prediction error of Perlin~$P_2$ and $P_3$, as shown in \cref{fig:p2} and \cref{fig:p3}, the error, again in terms of mean/median, increases less severe than altered training data evaluation would suggest. In general, for both cases the statistical magnitude of the induced error during training is greatly reduced with regard to the model's performance. Furthermore, the difference in mAP as in \cref{table_map} suggests a much steeper difference in performance, a property justifiably with the threshold-based definition of mAP.

\section{Conclusions}

Based on real-world problem cases, we model three different localization noise scenarios, Ramp, Gaussian and Perlin noise, which we apply to the geographical training split of the Argoverse 2 dataset. We train individual models based on the MapTRv2 architecture for a set of different pragmatically chosen noise parameters.

Besides the common mAP metric and qualitative analysis we introduce a novel detection to ego pose distance-aware metric, that allows for a more structured analysis of our altered training labels as well as model predictions, which we evaluate against the validation splits of unaltered ground truth.

Here, we show the stronger reliance on a low angular error compared to a translational error for adequate map perception performance, especially at higher distances from the ego vehicle.
This angular error is applied directly in some of your noise cases, in others it is created by the heading correction. This we employ to correct the changed direction of travel caused by our introduced translational error.
The negative consequence of heading correction on the model's performance cannot be understated at this point and should be considered during localization filter design.

Since one of our localization noise patterns, Gaussian noise, does not produce a temporally consistent localization error, we show that, based on otherwise statistically similar label input, the models perform very comparable. Models with temporal dependencies in their architecture may react differently though.

By altering only a percentage of the ground truth, we show the model's performance to be enhanced beyond what this percentage would infer by the presence of correct labels. Having few bad sections in an otherwise good localization during dataset recording should therefore still yield good model performance. 

Through our qualitative analysis, we deliver an assessment regarding the level of localization noise applied and the model no longer producing adequate map perception performance. Whereas smaller levels of localization errors can be largely compensated by the model, medium levels still produce logically coherent results albeit with increased levels of warping. In a multi-drive fleet data scenario, it should still be possible to compensate for these effects through aggregation of multiple drives. At high levels of localization error, the overall structure of the scene gets lost; such scenarios should be avoided.

\section*{Acknowledgments}
This work results from the just better DATA (jbDATA) project supported by the German Federal Ministry for Economic Affairs and Climate Action of Germany (BMWK) and the European Union, grant number 19A23003H and was supported by Helmholtz AI computing resources (HAICORE) of the Helmholtz Association’s Initiative and Networking Fund through Helmholtz AI.

\printbibliography

@misc{immel2024m3trgeneralisthdmap,
      title={M3TR: Generalist HD Map Construction with Variable Map Priors}, 
      author={Fabian Immel and Richard Fehler and Frank Bieder and Jan-Hendrik Pauls and Christoph Stiller},
      year={2024},
      eprint={2411.10316},
      archivePrefix={arXiv},
      primaryClass={cs.CV},
      url={https://arxiv.org/abs/2411.10316}, 
}

@inproceedings{carion_2020_detr,
	address = {Cham},
	title = {End-to-{End} {Object} {Detection} with {Transformers}},
	isbn = {978-3-030-58452-8},
	booktitle = {Computer {Vision} – {ECCV} 2020},
	publisher = {Springer International Publishing},
	author = {Carion, Nicolas and Massa, Francisco and Synnaeve, Gabriel and Usunier, Nicolas and Kirillov, Alexander and Zagoruyko, Sergey},
	editor = {Vedaldi, Andrea and Bischof, Horst and Brox, Thomas and Frahm, Jan-Michael},
	year = {2020},
	pages = {213--229},
}

@inproceedings{liu2023vectormapnet,
author = {Liu, Yicheng and Yuan, Tianyuan and Wang, Yue and Wang, Yilun and Zhao, Hang},
title = {VectorMapNet: end-to-end vectorized HD map learning},
year = {2023},
publisher = {JMLR.org},
articleno = {930},
numpages = {18},
location = {Honolulu, Hawaii, USA},
series = {ICML'23}
}

@inproceedings{liao2023MapTR,
  title={MapTR: Structured Modeling and Learning for Online Vectorized HD Map Construction},
  author={Liao, Bencheng and Chen, Shaoyu and Wang, Xinggang and Cheng, Tianheng, and Zhang, Qian and Liu, Wenyu and Huang, Chang},
  booktitle={International Conference on Learning Representations},
  year={2023}
}

@article{liao2024maptrv2,
  title={Maptrv2: An end-to-end framework for online vectorized hd map construction},
  author={Liao, Bencheng and Chen, Shaoyu and Zhang, Yunchi and Jiang, Bo and Zhang, Qian and Liu, Wenyu and Huang, Chang and Wang, Xinggang},
  journal={International Journal of Computer Vision},
  pages={1--23},
  year={2024},
  publisher={Springer}
}

@inproceedings{chen_maptracker_2025,
	address = {Cham},
	title = {{MapTracker}: {Tracking} with {Strided} {Memory} {Fusion} for {Consistent} {Vector} {HD} {Mapping}},
	isbn = {978-3-031-72658-3},
	booktitle = {Computer {Vision} – {ECCV} 2024},
	publisher = {Springer Nature Switzerland},
	author = {Chen, Jiacheng and Wu, Yuefan and Tan, Jiaqi and Ma, Hang and Furukawa, Yasutaka},
	editor = {Leonardis, Aleš and Ricci, Elisa and Roth, Stefan and Russakovsky, Olga and Sattler, Torsten and Varol, Gül},
	year = {2025},
	pages = {90--107},
}

@inproceedings{wang2024StreamQueryDenoising,
author = {Wang, Shuo and Jia, Fan and Mao, Weixin and Liu, Yingfei and Zhao, Yucheng and Chen, Zehui and Wang, Tiancai and Zhang, Chi and Zhang, Xiangyu and Zhao, Feng},
title = {Stream Query Denoising for Vectorized HD-Map Construction},
year = {2024},
isbn = {978-3-031-72654-5},
publisher = {Springer-Verlag},
address = {Berlin, Heidelberg},
url = {https://doi.org/10.1007/978-3-031-72655-2_12},
doi = {10.1007/978-3-031-72655-2_12},
booktitle = {Computer Vision – ECCV 2024: 18th European Conference, Milan, Italy, September 29–October 4, 2024, Proceedings, Part XIX},
pages = {203–220},
numpages = {18},
location = {Milan, Italy}
}

@InProceedings{Yuan_2024_WACV,
    author    = {Yuan, Tianyuan and Liu, Yicheng and Wang, Yue and Wang, Yilun and Zhao, Hang},
    title     = {StreamMapNet: Streaming Mapping Network for Vectorized Online HD Map Construction},
    booktitle = {Proceedings of the IEEE/CVF Winter Conference on Applications of Computer Vision (WACV)},
    month     = {January},
    year      = {2024},
    pages     = {7356-7365}
}

@INPROCEEDINGS{luo2024smerf,
  author={Luo, Katie Z and Weng, Xinshuo and Wang, Yan and Wu, Shuang and Li, Jie and Weinberger, Kilian Q and Wang, Yue and Pavone, Marco},
  booktitle={2024 IEEE International Conference on Robotics and Automation (ICRA)}, 
  title={Augmenting Lane Perception and Topology Understanding with Standard Definition Navigation Maps}, 
  year={2024},
  volume={},
  number={},
  pages={4029-4035},
  doi={10.1109/ICRA57147.2024.10610276}
}

@misc{zeng2024drivingpriormapsunified,
      title={Driving with Prior Maps: Unified Vector Prior Encoding for Autonomous Vehicle Mapping}, 
      author={Shuang Zeng and Xinyuan Chang and Xinran Liu and Zheng Pan and Xing Wei},
      year={2024},
      eprint={2409.05352},
      archivePrefix={arXiv},
      primaryClass={cs.CV},
      url={https://arxiv.org/abs/2409.05352}, 
}

@inproceedings{northcutt2021labelerrors,
      title={Pervasive Label Errors in Test Sets Destabilize Machine Learning Benchmarks}, 
      author={Curtis G. Northcutt and Anish Athalye and Jonas Mueller},
      month={December},
      year={2021},
      booktitle={Proceedings of the 35th Conference on Neural Information Processing Systems Track on Datasets and Benchmarks}
}

@InProceedings{Li_2020_CVPR,
author = {Li, Hengduo and Wu, Zuxuan and Zhu, Chen and Xiong, Caiming and Socher, Richard and Davis, Larry S.},
title = {Learning From Noisy Anchors for One-Stage Object Detection},
booktitle = {Proceedings of the IEEE/CVF Conference on Computer Vision and Pattern Recognition (CVPR)},
month = {June},
year = {2020}
}

@InProceedings{Chen_2023_ICCV,
    author    = {Chen, Zehui and Li, Zhenyu and Wang, Shuo and Fu, Dengpan and Zhao, Feng},
    title     = {Learning from Noisy Data for Semi-Supervised 3D Object Detection},
    booktitle = {Proceedings of the IEEE/CVF International Conference on Computer Vision (ICCV)},
    month     = {October},
    year      = {2023},
    pages     = {6929-6939}
}

@misc{zhu2024robusttinyobjectdetection,
      title={Robust Tiny Object Detection in Aerial Images amidst Label Noise}, 
      author={Haoran Zhu and Chang Xu and Wen Yang and Ruixiang Zhang and Yan Zhang and Gui-Song Xia},
      year={2024},
      eprint={2401.08056},
      archivePrefix={arXiv},
      primaryClass={cs.CV},
      url={https://arxiv.org/abs/2401.08056}, 
}

@article{freire_beyond_2024,
	title = {Beyond clean data: {Exploring} the effects of label noise on object detection performance},
	volume = {304},
	issn = {0950-7051},
	url = {https://www.sciencedirect.com/science/article/pii/S095070512401178X},
	doi = {https://doi.org/10.1016/j.knosys.2024.112544},
	journal = {Knowledge-Based Systems},
	author = {Freire, Agostinho and Silva, Leandro H. de S. and Andrade, João V. R. de and Azevedo, George O. A. and Fernandes, Bruno J. T.},
	year = {2024},
	pages = {112544},
}

@inproceedings{adhikari_effect_2021,
	address = {Cham},
	title = {Effect of {Label} {Noise} on {Robustness} of {Deep} {Neural} {Network} {Object} {Detectors}},
	isbn = {978-3-030-83906-2},
	booktitle = {Computer {Safety}, {Reliability}, and {Security}. {SAFECOMP} 2021 {Workshops}},
	publisher = {Springer International Publishing},
	author = {Adhikari, Bishwo and Peltomäki, Jukka and Germi, Saeed Bakhshi and Rahtu, Esa and Huttunen, Heikki},
	editor = {Habli, Ibrahim and Sujan, Mark and Gerasimou, Simos and Schoitsch, Erwin and Bitsch, Friedemann},
	year = {2021},
	pages = {239--250},
}

@article{hao2024mapbench,
    author = {Xiaoshuai Hao and Mengchuan Wei and Yifan Yang and Haimei Zhao and Hui Zhang and Yi Zhou and Qiang Wang and Weiming Li and Lingdong Kong and Jing Zhang},
    title = {Is Your HD Map Constructor Reliable under Sensor Corruptions?},
    journal={arXiv preprint arXiv:2406.12214},
    year = {2024},
}

@INPROCEEDINGS{Beemelmanns_iv_2024,
  author={Beemelmanns, Till and Zhang, Quan and Geller, Christian and Eckstein, Lutz},
  booktitle={2024 IEEE Intelligent Vehicles Symposium (IV)}, 
  title={MultiCorrupt: A Multi-Modal Robustness Dataset and Benchmark of LiDAR-Camera Fusion for 3D Object Detection}, 
  year={2024},
  volume={},
  number={},
  pages={3255-3261},
  doi={10.1109/IV55156.2024.10588664}
}

@inproceedings{perlin_noise_2002,
author = {Perlin, Ken},
title = {Improving noise},
year = {2002},
isbn = {1581135211},
publisher = {Association for Computing Machinery},
address = {New York, NY, USA},
url = {https://doi.org/10.1145/566570.566636},
doi = {10.1145/566570.566636},
booktitle = {Proceedings of the 29th Annual Conference on Computer Graphics and Interactive Techniques},
pages = {681–682},
numpages = {2},
location = {San Antonio, Texas},
series = {SIGGRAPH '02}
}

@article{Kos2010EffectsOM,
  title={Effects of multipath reception on GPS positioning performance},
  author={Tomislav Kos and Ivan Marke{\v{z}}i{\'c} and Josip Pokrajcic},
  journal={Proceedings ELMAR-2010},
  year={2010},
  pages={399-402},
  url={https://api.semanticscholar.org/CorpusID:36084317}
}

@article{Riekert1940ZurFD,
  title={Zur Fahrmechanik des gummibereiften Kraftfahrzeugs},
  author={P. Riekert and Tobias Schunck},
  journal={Ingenieur-Archiv},
  year={1940},
  volume={11},
  pages={210-224},
  url={https://api.semanticscholar.org/CorpusID:198141217}
}

@INPROCEEDINGS { Argoverse2,
  author = {Benjamin Wilson and William Qi and Tanmay Agarwal and John Lambert and Jagjeet Singh and Siddhesh Khandelwal and Bowen Pan and Ratnesh Kumar and Andrew Hartnett and Jhony Kaesemodel Pontes and Deva Ramanan and Peter Carr and James Hays},
  title = {Argoverse 2: Next Generation Datasets for Self-driving Perception and Forecasting},
  booktitle = {Proceedings of the Neural Information Processing Systems Track on Datasets and Benchmarks (NeurIPS Datasets and Benchmarks 2021)},
  year = {2021}
}

@inproceedings{caesar_nuscenes
,
  author       = {Holger Caesar and
                  Varun Bankiti and
                  Alex H. Lang and
                  Sourabh Vora and
                  Venice Erin Liong and
                  Qiang Xu and
                  Anush Krishnan and
                  Yu Pan and
                  Giancarlo Baldan and
                  Oscar Beijbom},
  title        = {nuScenes: {A} Multimodal Dataset for Autonomous Driving},
  booktitle    = {2020 {IEEE/CVF} Conference on Computer Vision and Pattern Recognition,
                  {CVPR} 2020, Seattle, WA, USA, June 13-19, 2020},
  pages        = {11618--11628},
  publisher    = {Computer Vision Foundation / {IEEE}},
  year         = {2020},
  url          = {https://openaccess.thecvf.com/content\_CVPR\_2020/html/Caesar\_nuScenes\_A\_Multimodal\_Dataset\_for\_Autonomous\_Driving\_CVPR\_2020\_paper.html},
  doi          = {10.1109/CVPR42600.2020.01164},
  bibsource    = {dblp computer science bibliography, https://dblp.org}
}

@InProceedings{Lilja_2024_CVPR,
    author    = {Lilja, Adam and Fu, Junsheng and Stenborg, Erik and Hammarstrand, Lars},
    title     = {Localization Is All You Evaluate: Data Leakage in Online Mapping Datasets and How to Fix It},
    booktitle = {Proceedings of the IEEE/CVF Conference on Computer Vision and Pattern Recognition (CVPR)},
    month     = {June},
    year      = {2024},
    pages     = {22150-22159}
}

@inproceedings{resnet,
author = {He, Kaiming and Zhang, Xiangyu and Ren, Shaoqing and Sun, Jian},
booktitle = {Proceedings of the IEEE Conference on Computer Vision and Pattern Recognition (CVPR)},
year = {2016},
month = {06},
pages = {770-778},
title = {Deep Residual Learning for Image Recognition},
doi = {10.1109/CVPR.2016.90}
}

\end{document}